\RequirePackage{float}
\documentclass[a4paper,fleqn]{cas-dc}
\ExplSyntaxOn
\keys_set:nn { stm / mktitle } { nologo = true }
\ExplSyntaxOff
\usepackage[authoryear,sort]{natbib}
\usepackage{enumitem}
\usepackage{algorithm}
\makeatletter
\renewcommand{\floatc@ruled}[2]{{\normalfont\bfseries #1.}\enspace #2\par}
\makeatother
\usepackage{xurl}
\input{citation_commands}
\bibpunct{(}{)}{;}{a}{,}{,}
\shorttitle{Instance-preserving graded semantic reporting}
\shortauthors{Xue et al.}
\title[mode=title]{Instance-preserving graded semantic reporting for surgical action triplets}
\begin{document}
\let\printorcid\relax
\author[a]{Yanchong Xue}
\ead{thouofsgs@hrbeu.edu.cn}

\author[a]{Wanting Zhang}
\ead{zhangwt@hrbeu.edu.cn}

\author[a]{Lingtao Yu}
\cormark[1]
\ead{yulingtaohrbeu@163.com}

\author[b]{Guochao Zhang}
\ead{m18689856730@163.com}

\affiliation[a]{organization={College of Mechanical and Electrical Engineering, Harbin Engineering University},
    addressline={No. 145 Nantong Street, Nangang District},
    city={Harbin}, postcode={150001}, state={Heilongjiang Province}, country={P. R. China}}

\affiliation[b]{organization={Department of Thoracic Surgery I, Hainan Cancer Hospital},
    addressline={No. 6 Changbin West 4th Street, Xiuying District},
    city={Haikou}, postcode={570312}, state={Hainan Province}, country={P. R. China}}

\cortext[1]{Corresponding author.}

\begin{abstract}
Complete-or-reject reporting of surgical action triplets can discard a correctly identified instrument and action when the target is uncertain. We introduce instance-preserving graded semantic reporting, which separates instance selection from the choice of semantic detail. Each selected instance retains its predicted mask and labels while hierarchical correctness scores guide instrument (I), instrument-verb (IV), or complete instrument-verb-target (IVT) reporting. The utility-increment rule DELTA weighs the expected gain from finer reporting against its error cost. We compare three decision rules on 5,641 test frames from eight surgical videos using fixed predictions and scores. At the 10\% development calibration target, DELTA retains correct partial reports for 80.0\% of candidates rejected by a matched complete-or-reject baseline (5,329/6,662). Adding IV increases source-normalized weighted correct-report yield by 19.13 to 23.13 per 100 frames across all three rules relative to separately recalibrated I/IVT controls. Weighted reporting risk changes by -0.77 to +1.01 percentage points, with complete correct-triplet counts unchanged or reduced. Graded reporting makes partial surgical predictions available as localized descriptions, with an explicit trade-off between report utility and completeness.
\end{abstract}
\maketitle
\ifdefined\XeTeXversion
  
\else
  \ifdefined\pdfminorversion\pdfminorversion=7\fi
\fi
\hypersetup{pdfauthor={Yanchong Xue, Wanting Zhang, Lingtao Yu, Guochao Zhang},pdftitle={Instance-preserving graded semantic reporting for surgical action triplets},pdfsubject={}}
\section{Introduction}\label{sec:1}

\begin{figure}[pos=t]
\centering
\includegraphics{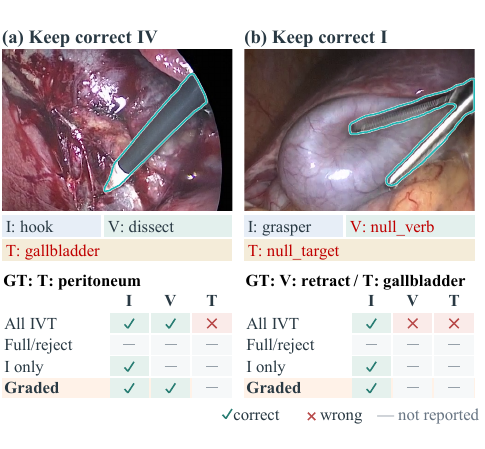}
\caption{Partial semantics retained by graded reporting on two surgical instances. (a) The predicted hook and dissect action are correct, while the target is gallbladder instead of peritoneum; the graded report retains IV. (b) The grasper is correct, while the predicted verb and target are incorrect; the graded report retains I. Cyan contours show the predicted instrument masks. Predictions are fixed across the output choices within each example. All IVT and I only display fixed semantic levels; Full/reject denotes complete-or-reject reporting. Full/reject and Graded use the main configurations fixed before evaluation, with Graded illustrating the utility-increment rule DELTA. Check marks and crosses indicate field correctness assessed against ground truth; horizontal strokes indicate omitted fields. The null\_verb and null\_target entries are predicted class labels; ground-truth (GT) rows show fields that differ from the predictions. I, IV and IVT denote instrument, instrument-verb and full triplet reports, respectively.}\label{fig:1}
\end{figure}

Surgical action triplets describe an interaction through the instrument, the action it performs and its anatomical target \citep{ref01}. Triplet detection adds spatial localization \citep{ref02}, while triplet segmentation links each interaction to a pixel-level instrument mask, as formalized in CholecTriplet-Seg \citep{ref03}. This association makes it possible to distinguish concurrent actions and relate semantic predictions to visible instruments. Yet the fields attached to an instance can differ in reliability. In Fig.~\ref{fig:1}(a), a hook and its dissecting action are correctly identified, whereas the predicted target is incorrect. The instrument-action pair still describes part of the observed interaction. In Fig.~\ref{fig:1}(b), the instrument alone is correctly recognized. These examples motivate a reporting problem: deciding how much semantic detail to attach to a selected spatial instance.

Choosing a report determines which predicted fields the recipient sees. Reporting every complete triplet exposes all predicted fields, including uncertain verbs or targets. A complete-or-reject policy can withhold a prediction with low confidence in its full triplet, taking its usable components out of the report as well. Instrument-only output retains the selected instances but omits their actions even when those actions are reliable. An intermediate instrument-verb report offers a useful alternative when target identity is uncertain. Choosing among these outputs calls for separate decisions about which instances enter the report and how much detail each instance carries.

Selective prediction provides an established foundation for choosing which outputs to report \citep{ref04}. Confidence-based rejection allows a trained classifier to trade coverage for accuracy \citep{ref05,ref06,ref07}. Hierarchical recognition extends the available choices by allowing a prediction to move to a broader semantic category \citep{ref08}. \citet{ref09} formulated this accuracy-specificity trade-off and introduced Dual Accuracy Reward Trade-off Search (DARTS). \citet{ref10} developed hierarchical selective classification, including inference rules and hierarchical risk-coverage evaluation. These studies establish that uncertainty can be expressed through the specificity of an output \citep{ref11}. For surgical action reports, we consider an instrument (I), an instrument-verb pair (IV) and a complete instrument-verb-target triplet (IVT), each attached to the same predicted instrument mask.

Spatially grounded reports bring an additional requirement: semantic correctness must remain tied to a particular instance \citep{ref12}. A report can fail through incorrect localization, competition for the same instrument, or an error in a displayed field. Its evaluation therefore depends on both the selected report set and the fields shown within each report. Changing the retained set can alter matching and correct-report yield before semantic granularity is considered. Holding that set fixed allows the contribution of semantic detail to be examined on the same spatial instances.

We introduce instance-preserving graded semantic reporting to select the semantic detail attached to each admitted instrument instance. A base selection step establishes the candidate set and its ordering. A subsequent reporting step chooses I, IV or IVT for each selected candidate, retaining its mask and the predicted labels of its displayed fields. The framework uses hierarchical correctness scores to decide when additional semantic detail is worth reporting. We examine threshold-based selection (THR), an adaptation of DARTS, and the utility-increment rule DELTA under this shared interface. DELTA evaluates the expected gain from a finer report against its additional error cost, using I as the reference.

The intermediate IV level lets the report retain an action description when confidence in the complete triplet is low. We compare three-level reporting against separately recalibrated I/IVT controls, sharing predictions, correctness scores and the same admitted instances (base support). At the 10\% development calibration target, adding IV increases weighted correct-report yield by 19.13 to 23.13 per 100 frames on the test set. Yield is normalized by video source; each rule also shows a positive gain in all eight evaluated videos. Comparisons with complete-or-reject output establish how partial reports retain usable semantics among rejected candidates. Instance-level analyses identify the retained semantics and how these choices affect complete correct-triplet output and observed reporting risk.

The study makes two contributions:

\begin{enumerate}
\item An instance-preserving reporting framework for surgical action triplets that separates base instance selection from semantic granularity, supports I, IV and IVT outputs, and evaluates their joint spatial and semantic correctness.
\item Evidence that partial reports retain usable semantics among rejected candidates and that the intermediate IV level increases correct-report yield across the evaluated rules. Recalibrated controls and instance-level analyses establish this contribution and explain its effects on complete records and reporting risk.
\end{enumerate}

Section~\ref{sec:2} reviews surgical triplet prediction and selective reporting. Section~\ref{sec:3} defines the framework, and Section~\ref{sec:4} describes the evaluation protocol. Section~\ref{sec:5} presents the results, followed by discussion and conclusions in Sections~\ref{sec:6} and~\ref{sec:7}.

\section{Related work}\label{sec:2}

\subsection{Surgical action triplets and instance grounding}\label{sec:2.1}

Triplet recognition models use the dependencies among instruments, verbs and targets to identify interactions within an image \citep{ref13,ref14,ref15,ref16}. Tripnet guides verb and target recognition with instrument activation maps and learns component associations through a trainable interaction space \citep{ref17}. Rendezvous develops attention mechanisms for component recognition and models their relationships through self- and cross-attention \citep{ref01}. These methods predict component labels and their combinations, providing structured descriptions of surgical activity.

Spatial grounding assigns these interactions to individual instruments \citep{ref18,ref19}. CholecTriplet2022 benchmarks triplet detection through instrument bounding boxes and their associated action labels \citep{ref02}. \citet{ref20} combine instrument-aware target embeddings with an interaction graph to associate localized instruments with verbs and targets. At the pixel level, \citet{ref03} introduce CholecTriplet-Seg and TargetFusionNet, which combines instance queries with weak anatomy priors to improve target prediction. The segmentation formulation couples an instrument mask with its action and target. Our reporting framework starts from such instance predictions and chooses the semantic fields displayed for each retained mask.

\subsection{Selective and hierarchical prediction}\label{sec:2.2}

Selective prediction treats acceptance as part of the prediction problem \citep{ref21}. Confidence-based selection trades the fraction of accepted samples against their error rate \citep{ref05,ref22}, while SelectiveNet jointly learns prediction and rejection \citep{ref23}. Confidence calibration addresses a related requirement: how closely predicted probabilities reflect empirical correctness \citep{ref24}. \citet{ref25} show that temperature scaling can improve this correspondence for neural classifiers. For structured reports, the event being scored must also be specified, since correctness depends on the combination of fields being displayed \citep{ref26,ref27}.

Hierarchical prediction adds choices at different levels of semantic specificity \citep{ref08}. \citet{ref09} formulate recognition as maximizing specificity reward subject to an accuracy requirement and introduce DARTS to choose among hierarchy nodes. \citet{ref10} formalize hierarchical selective risk and coverage and develop inference rules for moving from specific predictions to broader ancestors. Their Climbing rule follows the path from the most likely leaf towards the root until the confidence threshold is met. The selected hierarchy level determines how specific an accepted prediction is \citep{ref28}.

These approaches provide the basis for choosing I, IV or IVT on a fixed reporting path for each selected instrument instance. Under nested correctness scores and an instrument-level fallback, the threshold rule THR is equivalent to Climbing on this path. Our DARTS comparison adapts its decision rule to the same scores, report weights and development calibration procedure used by the other rules. The hierarchy here changes the fields displayed for an existing prediction, while its mask and labels remain fixed. The common reporting interface lets us compare semantic specificity while keeping spatial instance retention fixed.

\subsection{Evaluation of structured reports}\label{sec:2.3}

Triplet evaluation distinguishes individual components, combinations of fields and spatially grounded interactions. The ivtmetrics protocol reports average precision for individual components, instrument-verb and instrument-target pairs, and complete triplets \citep{ref29}. Component-level scores describe recognition of each semantic category; pair and triplet scores assess increasingly specific combinations \citep{ref30,ref31}. Triplet segmentation adds pixel-level grounding, requiring semantic correctness together with instrument localization \citep{ref03}. These evaluations characterize predicted content at predefined semantic levels. Graded reporting introduces a further choice: the fields displayed for one instance can differ from those displayed for another.

Selective risk-coverage evaluation measures error among accepted predictions as acceptance changes \citep{ref05,ref32,ref33}, while hierarchical coverage accounts for the specificity of selected nodes \citep{ref10}. A grounded report is correct when its mask matches a ground-truth instance uniquely and every displayed field agrees with that instance. An IV report can therefore be correct even when its full triplet would contain an incorrect target. Joint report correctness complements component metrics by retaining the association among the displayed fields and their spatial instance.

We measure weighted correct-report yield together with reporting risk and complete correct-triplet output. Comparisons account for the retained instance sets and use fixed predictions and scores to distinguish the effects of instance support, available report levels and decision rules. Section~\ref{sec:3} develops this formulation into an executable reporting framework.

\section{Problem formulation and reporting framework}\label{sec:3}

\subsection{Inputs, outputs and overview}\label{sec:3.1}

We formulate reporting as a choice of semantic detail for each selected instrument instance. Following the triplet segmentation setting \citep{ref03}, a frozen predictor associates an instrument mask with an instrument, verb and target. For a frame \(x\), let \(\mathcal{C}_{x}\) denote its candidate queries, the predictor's instance slots. Candidate \(j\) carries a binary mask \(M_{j}\), a predicted triplet \({\hat{t}}_{j} = ({\hat{I}}_{j},{\hat{V}}_{j},{\hat{T}}_{j})\) and a 101-way class-probability vector \(\boldsymbol{\pi}_{j}\). Its first 100 entries represent triplet classes, and the remaining entry represents no object. The fixed triplet label is the highest-probability class among the first 100 entries.

The candidate pool is formed from the 100 highest-probability query-class pairs over the 100 triplet classes. We retain distinct queries with nonempty masks. Mask logits are restored to the original image coordinates and thresholded at zero. A report retains its mask and displays the projection \(P_{a}({\hat{t}}_{j})\) for an action \(a \in \mathcal{A} = \{ \mathrm{I},\mathrm{IV},\mathrm{IVT}\}\). These projections contain the instrument, the instrument-verb pair, or the complete triplet. Displayed values retain their predicted labels. An omitted field is left blank; null\_verb and null\_target remain ordinary classes in the label dictionary.

\begin{figure*}[pos=!t]
\centering
\includegraphics{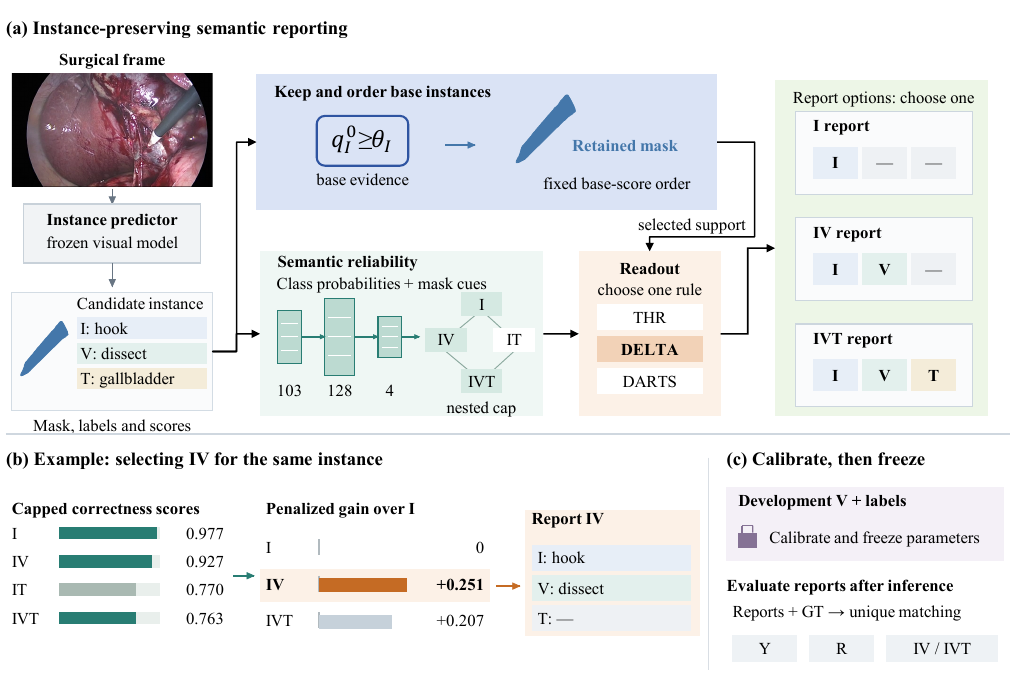}
\caption{Instance-preserving graded semantic reporting. (a) Base evidence selects and orders instrument instances; correctness scores guide a choice among three mutually exclusive reporting levels. Each report retains the candidate mask and the labels of its displayed fields. THR, DELTA and the DARTS adaptation share this interface. (b) A fixed hook/dissect/gallbladder prediction illustrates DELTA with an error-cost parameter of 1. The score-based gains over I are 0.251 for IV and 0.207 for IVT, so the output displays hook/dissect and omits the target. The instrument-target pair (IT) supplies an auxiliary correctness score for the nested cap. (c) Development V denotes the calibration data. Parameters are frozen before inference, and ground truth enters the subsequent matching and evaluation stage.}\label{fig:2}
\end{figure*}

The framework separates instance admission from semantic reporting, as shown in Fig.~\ref{fig:2}. The base instrument score determines which candidates enter the reporting stage and in what order. The correctness scorer supplies the evidence used to choose one reporting level for each admitted instance. Development calibration fixes the base threshold and rule parameters before evaluation. At inference, the frozen program returns one report for every admitted instance. Calibration and final evaluation use the matching procedure defined in Section~\ref{sec:3.5}. The reporting rules therefore vary displayed detail while preserving the admitted masks, predicted labels and ordering.

\subsection{Base instance support}\label{sec:3.2}

Instance admission uses evidence for the predicted instrument. Let \(m_{j}\) be mask quality, computed as the mean sigmoid mask probability over foreground pixels at the original image resolution. The base instrument score aggregates the unconditional triplet probabilities associated with \({\hat{I}}_{j}\) and multiplies this sum by mask quality. Writing \(I(k)\) for the instrument of triplet class \(k\), we define

\begin{equation}\label{eq:1}
\begin{aligned}
q_{j,\mathrm{I}}^{0} &= m_{j}\sum_{k:I(k) = {\hat{I}}_{j}}^{}\pi_{jk},\\ \mathcal{B}_{x} &= \{ j \in \mathcal{C}_{x}:q_{j,\mathrm{I}}^{0} \geq \theta_{\mathrm{I}}\}.
\end{aligned}
\end{equation}

Here \(\pi_{jk}\) is the probability assigned to triplet class \(k\), and \(\theta_{I}\) is the base admission threshold. The superscript 0 identifies the base score. The 101-class softmax normalization is retained when forming this sum. Within each frame, we order admitted candidates by decreasing \(q_{j,\mathrm{I}}^{0}\), breaking ties by increasing query identifier. All threshold-passing candidates remain in the main reporting policies. Their semantic choices therefore preserve the base instance set \(\mathcal{B}_{x}\) and its ordering.

Every available action contains the instrument field. If \(\mathcal{R}_{x}\) is the ordered collection of reports, its instrument projection is

\begin{equation}\label{eq:2}
P_{\mathrm{I}}(\mathcal{R}_{x}) = \lbrack(M_{j},{\hat{I}}_{j}):j \in \mathcal{B}_{x}\rbrack.
\end{equation}

The brackets retain the base ordering. Under the geometric matching rule in Section~\ref{sec:3.5}, the unchanged masks and ordering preserve the ground-truth assignments and correct instrument count. For policies with the same base threshold, this invariance separates changes in semantic output from changes in instrument retention. A different calibration target can select another base set, while complete-or-reject reporting filters that set further.

\subsection{Hierarchical correctness scoring}\label{sec:3.3}

The scorer learns joint correctness at four semantic levels. Training labels use a local geometric assignment: each candidate selects the ground-truth mask with the highest intersection over union (IoU), provided that IoU is at least \(\tau = 0.5\). Equal overlaps are resolved by ascending ground-truth identifier. This assignment is computed independently for each candidate, so several candidates may share a ground-truth instance during training. Let \(b_{j,f}^{\mathrm{loc}}\) indicate that an eligible local match exists and that the predicted field \(f\) agrees with that match. Unmatched candidates receive zero labels. The four supervision events are

\begin{equation}\label{eq:3}
\boldsymbol{z}_{j} = (b_{j,I}^{\mathrm{loc}},\ b_{j,I}^{\mathrm{loc}}b_{j,V}^{\mathrm{loc}},\ b_{j,I}^{\mathrm{loc}}b_{j,T}^{\mathrm{loc}},\ b_{j,I}^{\mathrm{loc}}b_{j,V}^{\mathrm{loc}}b_{j,T}^{\mathrm{loc}}).
\end{equation}

The input combines the class distribution with two mask cues. With \(\rho_{j}\) denoting mask area as a fraction of the image, the 103-dimensional feature vector and four raw scores are

\begin{equation}\label{eq:4}
\begin{aligned}
\boldsymbol{h}_{j} & = \lbrack\boldsymbol{\pi}_{j},m_{j},\log(1 + 100\rho_{j})\rbrack, \\
\boldsymbol{q}_{j} & = \sigma\left( W_{2}\,\operatorname{GELU}(W_{1}\boldsymbol{h}_{j} + \boldsymbol{b}_{1}) + \boldsymbol{b}_{2} \right).
\end{aligned}
\end{equation}

Here \(\sigma\) applies a sigmoid to each output logit. GELU denotes the Gaussian error linear unit activation; \(W_{1},W_{2}\) and \(\boldsymbol{b}_{1},\boldsymbol{b}_{2}\) are learned weight matrices and bias vectors. The network has 128 hidden units and four output units, ordered as I, IV, instrument-target (IT) and IVT, for a total of 13,828 trainable parameters. Scorer training uses three disjoint folds of video sources. For each held-out fold, a predictor fitted on the other two folds produces the training candidates. The resulting candidates and local labels are pooled to train the scorer.

Binary cross-entropy is averaged over the four events and then over candidates within each frame. Each nonempty frame from source \(s\) receives weight \(F/(|\mathcal{S}_{\mathrm{train}}|F_{s})\), where \(\mathcal{S}_{\mathrm{train}}\) is the set of training sources, and \(F_{s}\) and \(F\) count nonempty frames in source \(s\) and across all training sources, respectively. This gives each source equal total loss weight. Training uses AdamW \citep{ref34} with learning rate 0.001, weight decay 0.0001, batches of 32 frames and gradient-norm clipping at 1. The scorer is fixed after 20 epochs; Section~\ref{sec:4} specifies the source partitions and predictor training.

Before selecting a report, we obtain the capped correctness scores \(\boldsymbol{p}_{j}\) by enforcing event inclusion:

\begin{equation}\label{eq:5}
\begin{aligned}
p_{j,I} & = q_{j,I}, \\ p_{j,IV} & = \min(q_{j,IV},p_{j,I}), \\
p_{j,IT} & = \min(q_{j,IT},p_{j,I}), \\ p_{j,IVT} & = \min(q_{j,IVT},p_{j,IV},p_{j,IT}).
\end{aligned}
\end{equation}

Thus a complete-triplet score is bounded by both pair scores, and each pair score is bounded by the instrument score. IT contributes to this constraint; the output menu remains I, IV and IVT. The base score \(q_{j,I}^{0}\) determines admission and order, whereas \(\boldsymbol{p}_{j}\) guides the choice of displayed fields. The scores estimate correctness for each candidate under local assignment. Calibration and final evaluation then account for competition among overlapping candidates through unique matching.

\subsection{Reporting rules and frozen inference}\label{sec:3.4}

The three rules use the same capped scores and base ordering. Report weights follow the fraction of displayed fields: \(w_{I} = 1/3\), \(w_{IV} = 2/3\) and \(w_{IVT} = 1\). These weights value a jointly correct report according to its semantic detail. We write \(a_{j}\) for the action selected for candidate \(j \in \mathcal{B}_{x}\).

THR selects the most specific level passing a common semantic threshold \(\theta\), with I available as the fallback:

\begin{equation}\label{eq:6}
a_{j}^{\mathrm{THR}} = \begin{cases}
\mathrm{IVT}, & p_{j,IVT} \geq \theta, \\
\mathrm{IV}, & p_{j,IV} \geq \theta > p_{j,IVT}, \\
\mathrm{I}, & \text{otherwise}.
\end{cases}
\end{equation}

Under the nested scores, this is the Climbing decision \citep{ref10} restricted to the fixed report path and its instrument-level fallback. The separate base threshold controls instance admission, while the semantic threshold controls refinement.

DELTA asks whether a finer report offers enough additional correct-report utility to justify its additional error cost. For each action, \(w_{a}p_{j,a}\) and \(w_{a}(1 - p_{j,a})\) provide score-based estimates of utility and error cost. Relative to I, the penalized gain is

\begin{equation}\label{eq:7}
\begin{aligned}
\Delta_{j,a} &= \left\lbrack w_{a}p_{j,a} - w_{I}p_{j,I} \right\rbrack\\ &\quad-\lambda\left\lbrack w_{a}(1 - p_{j,a}) - w_{I}(1 - p_{j,I}) \right\rbrack,\\ &\quad\lambda \geq 0.
\end{aligned}
\end{equation}

The parameter \(\lambda\) controls the penalty on additional incorrect-report weight. We compare IV and IVT directly against the same I reference and select the largest positive gain. If neither improves on zero by more than \(\varepsilon = 10^{- 12}\), the report remains I. Objective differences within this tolerance favor the coarser report. Figure~\ref{fig:2}(b) illustrates how IV can yield a larger gain than a complete triplet for the same candidate.

The DARTS adaptation follows the reward-accuracy decision of \citet{ref09}, using the present report weights and correctness scores:

\begin{equation}\label{eq:8}
a_{j}^{\mathrm{DARTS}} = \underset{a \in \mathcal{A}}{\arg\,\max}(w_{a} + \eta)p_{j,a},\text{\quad\quad}\eta \geq 0.
\end{equation}

The parameter \(\eta\) adds a reward for correctness to the weighted report utility. The rule uses the same tolerance and coarser-action tie convention. Both utility-based rules also include a designated I-only endpoint in their parameter grids. The shared interface permits each rule to operate on the restricted I/IVT menu by removing IV from its available choices.

For each empirical risk target, development calibration searches the candidate base thresholds and rule-specific parameters. It maximizes report utility subject to the target, using the unique matching and source normalization defined in Section~\ref{sec:3.5}. At the evaluated targets, the three calibrated rules selected the same base threshold. Matched controls use this base support: all-I and all-IVT fix the displayed level, and complete-or-reject retains IVT reports above its calibrated full-triplet threshold. Restricted I/IVT controls recalibrate their semantic parameters with the base threshold fixed. Section~\ref{sec:4} specifies the parameter grids and deterministic selection procedure. Algorithm~\ref{alg:1} summarizes inference with the selected parameters held fixed.

\begin{algorithm}[!t]
\caption{Frozen inference for graded semantic reporting}\label{alg:1}
\begin{minipage}{\linewidth}
\normalfont\normalsize
\textbf{Inputs:} Candidates \(\mathcal{C}_{x}\); trained logit network \(f\) from Eq.~\eqref{eq:4}; frozen \(\theta_{I}\), rule \(r\) and the corresponding semantic parameter \(\psi\).\par
\smallskip
\textbf{Output:} Reports \(\mathcal{R}_{x}\) in base-score order. Here \(\psi\) is \(\theta\), \(\lambda\) or \(\eta\) for THR, DELTA or DARTS, respectively.
\begin{enumerate}[leftmargin=1.8em,itemsep=0pt,parsep=0pt,topsep=4pt]
\item Compute \(q_{j,I}^{0}\) for each \(j \in \mathcal{C}_{x}\) using Eq.~\eqref{eq:1}.
\item \(\mathcal{B}_{x} \leftarrow \{ j:q_{j,I}^{0} \geq \theta_{I}\}\).
\item Order \(\mathcal{B}_{x}\) by \(\left( - q_{j,I}^{0},{\mathrm{query\_ id}}_{j} \right)\).
\item \(\mathcal{R}_{x} \leftarrow \lbrack\ \rbrack\).
\item For each \(j\) in the ordered \(\mathcal{B}_{x}\):
\item \hspace*{1em}Form \(\boldsymbol{h}_{j}\) using Eq.~\eqref{eq:4}.
\item \hspace*{1em}Compute \(\boldsymbol{q}_{j} = \sigma(f(\boldsymbol{h}_{j}))\).
\item \hspace*{1em}Apply the nested cap in Eq.~\eqref{eq:5}.
\item \hspace*{1em}\(a_{j} \leftarrow \operatorname{Select}(r,\boldsymbol{p}_{j};\psi)\).
\item \hspace*{1em}Append \(\left( M_{j},P_{a_{j}}({\hat{t}}_{j}) \right)\) to \(\mathcal{R}_{x}\).
\item End for.
\item Return \(\mathcal{R}_{x}\).
\end{enumerate}
\end{minipage}
\end{algorithm}

\subsection{Instance matching and report utility}\label{sec:3.5}

Final correctness is evaluated on the actual report set. Within each frame, candidates are visited in base-score order. Each receives the highest-IoU unused ground-truth instance satisfying \(\mathrm{IoU} \geq 0.5\); equal overlaps are resolved by the lexicographic ground-truth identifier. The assigned instance is then marked occupied, including when a displayed semantic label is wrong. A candidate with no eligible unused match remains unmatched. Complete-or-reject outputs undergo this matching again after filtering, so an instance released by rejection can be assigned to a later report.

Let \(o(j)\) be the assigned ground-truth instance and \(F(a_{j})\) the displayed fields. With \({\hat{y}}_{j,f}\) and \(y_{o(j),f}\) denoting the predicted and ground-truth labels for field \(f\), a report is correct when its geometry is matched and every displayed field agrees with that instance:

\begin{equation}\label{eq:9}
c_{j} = \begin{cases}
\prod_{f \in F(a_{j})}^{}{\mathbf{1}\lbrack}{\hat{y}}_{j,f} = y_{o(j),f}\rbrack, & o(j) \neq \varnothing, \\
0, & o(j) = \varnothing.
\end{cases}
\end{equation}

In Eq.~\eqref{eq:9}, the indicator \(\mathbf{1}\lbrack \cdot \rbrack\) equals one when its condition holds and zero otherwise; \(\varnothing\) denotes an unmatched report. To count correct displayed components, let \(b_{j,f}\) indicate a correct field under this same assignment, with zero for unmatched candidates. We record correct instrument projections, correct displayed IV pairs and complete correct triplets as

\begin{equation}\label{eq:10}
\begin{aligned}
C_{I} & = \sum_{j}^{}b_{j,I}, \\
C_{IV} & = \sum_{j}^{}{\mathbf{1}\lbrack}a_{j} \in \{ \mathrm{IV},\mathrm{IVT}\}\rbrack b_{j,I}b_{j,V}, \\
C_{IVT} & = \sum_{j}^{}{\mathbf{1}\lbrack}a_{j} = \mathrm{IVT}\rbrack c_{j}.
\end{aligned}
\end{equation}

These sums run over emitted reports. The IV count includes correct instrument-verb pairs displayed within IVT reports, even when their target is wrong. Equation~\eqref{eq:9} still scores that full report as incorrect. Recording both quantities preserves the distinction between useful displayed pairs and complete correct records.

We normalize reporting utility by video source. Let \(N_{s} > 0\) be the total number of evaluated frames in source \(s\), including frames with no reports, and \(\mathcal{R}_{s}\) its emitted reports. Define the displayed weight and jointly correct weight per frame as

\begin{equation}\label{eq:11}
\begin{aligned}
S_{s} &= \frac{1}{N_{s}}\sum_{j \in \mathcal{R}_{s}}^{}w_{a_{j}},\\
T_{s} &= \frac{1}{N_{s}}\sum_{j \in \mathcal{R}_{s}}^{}w_{a_{j}}c_{j}.
\end{aligned}
\end{equation}

For the fixed set of evaluation sources \(\mathcal{S}\), report utility and weighted reporting risk are

\begin{equation}\label{eq:12}
\begin{aligned}
Y_{\mathrm{info}} &= \frac{100}{\left| \mathcal{S} \right|}\sum_{s \in \mathcal{S}}^{}T_{s},\\
R_{\mathrm{info}} &= \frac{\sum_{s \in \mathcal{S}}^{}(S_{s} - T_{s})}{\sum_{s \in \mathcal{S}}^{}S_{s}}.
\end{aligned}
\end{equation}

Yield measures source-normalized weighted correct reports per 100 frames. Each source contributes equally to this mean; multiple instances per frame allow yield to exceed 100. Risk is the fraction of normalized displayed weight assigned to incorrect reports. A source with no output contributes zero displayed and correct weight and remains in the yield average. When total displayed weight is zero, yield is zero and risk is undefined.

For a matched prediction with correct I and V but incorrect T, an IV report contributes \(\frac{2}{3}\) correct weight and a full IVT report contributes zero. Coarsening a fully correct IVT report to IV changes its credit from \(1\) to \(\frac{2}{3}\). For nonempty output, \(Y_{\mathrm{info}} = 100(1 - R_{\mathrm{info}})\overline{S}\), where \(\overline{S}\) is the mean displayed weight per frame across sources.

\section{Experimental protocol}\label{sec:4}

The experiments tested two consequences of choosing semantic granularity on fixed predictions. Complete-or-reject comparisons measured correct partial descriptions among rejected candidates. Recalibrated I/IVT controls measured the value of IV on already retained instances. All comparisons used the predictor, correctness scorer and joint evaluation defined in Section~\ref{sec:3}.

\subsection{Data and predictor training}\label{sec:4.1}

We used CholecTriplet-Seg \citep{ref03}, which associates surgical action triplets with instrument instance masks. Table~\ref{tab:1} summarizes the video-disjoint populations used in this study. The training split was divided into D, V and C, containing 29, five and five videos, respectively. D supplied predictor training and the candidates used to learn correctness scores. V supported predictor checkpoint selection and reporting-parameter calibration; C supplied auxiliary evaluation and later analyses of stored model outputs. The main test population contained 5,641 frames from eight videos. The three-video val population supplied the later evaluation of unchanged frozen configurations. Here val names the transfer population, whereas V denotes development data.

\begin{table*}[pos=!t]
\caption{Data populations and their roles in the reporting study.}\label{tab:1}
\centering
\begin{tabular*}{\linewidth}{@{\extracolsep{\fill}}@{}lrrr>{\raggedright\arraybackslash}p{72mm}@{}}
\toprule
\textbf{Role} & \textbf{Videos} & \textbf{Frames} & \textbf{GT instances} & \textbf{Use} \\
\midrule
D & 29 & 16,160 & 27,012 & Predictor/scorer training \\
V & 5 & 4,157 & 7,235 & Parameter calibration \\
C & 5 & 2,118 & 3,344 & Cache diagnostics \\
test & 8 & 5,641 & 8,934 & Frozen test; cache analysis \\
val & 3 & 2,443 & 3,341 & Source transfer \\
\bottomrule
\end{tabular*}
\par\smallskip
{\sffamily\fontsize{8}{9.5}\selectfont\noindent D, V and C are video-disjoint subsets of the training split. Counts include every evaluated frame, including frames without output. Test comprises the original frozen evaluation and later cached-output comparisons; val is used for subsequent source transfer. GT denotes ground truth.\par}
\end{table*}

The visual predictor was a local Mask2Former adaptation \citep{ref35} with a ResNet50 encoder (a 50-layer residual network) \citep{ref36} and a joint classifier over 100 legal triplets plus no object. Its 100 queries produced instrument masks and fixed triplet labels. The encoder used ImageNet initialization \citep{ref37}, with batch-normalization \citep{ref38} parameters and statistics held fixed. Images were resized to a 640-pixel long edge while preserving aspect ratio. They were normalized with ImageNet statistics and padded to multiples of 32. Training applied horizontal flipping with probability 0.5. AdamW used an effective batch size of eight, learning rates of \(10^{- 4}\) for the decoder and \(10^{- 5}\) for the encoder, and weight decay 0.05. Normalization parameters and embeddings were exempt from weight decay.

The predictor trained on all of D completed 30 epochs with seed 20260915. Learning rates used 500 warmup updates followed by cosine decay towards 0.01 of their initial values over a fixed 50-epoch schedule. Checkpoints were evaluated on V at epochs 2, 4, 6, 8, 10, 12, 15, 18, 21, 24, 27 and 30. Joint triplet mask average precision selected epoch 24; exact ties favored the earlier checkpoint. This predictor then remained fixed throughout the reporting comparisons.

Scorer training used out-of-fold predictions from D, as described in Section~\ref{sec:3.3}. The three held-out folds contained 10, 10 and nine videos; their predictors were trained on the complementary 19, 19 and 20 videos. Each fold predictor used the same architecture, preprocessing, optimizer, effective batch size and learning-rate schedule. Epoch 24 was fixed as its endpoint without checkpoint search. The pooled cache contained 810,181 candidates over 16,160 frames, of which 15,926 had nonempty candidate sets. These candidates formed the training data for the correctness scorer with 103 input features, whose parameters were fixed after 20 epochs. V was then used to calibrate reporting decisions with both networks frozen.

\subsection{Reporting policies and matched controls}\label{sec:4.2}

We compared three kinds of output choice on shared predictions. All-I and all-IVT references displayed a fixed semantic level for every admitted instance. The matched complete-or-reject policy retained candidates whose capped full-triplet score passed a threshold calibrated on V. Graded policies selected I, IV or IVT while retaining every admitted instance. Complete-or-reject filtering changed the report set, so unique geometric matching was recomputed after filtering. Its initial candidate eligibility and base ordering matched those of the graded policies.

To isolate the intermediate IV level, we restricted each decision rule to the I/IVT menu and recalibrated its semantic parameter on V. The corresponding three-level and two-level policies shared the base threshold, masks, labels, correctness scores and ordering. Their difference was the available reporting menu and its calibrated semantic parameter. Within the same three-level interface, we also compared THR, DELTA and the DARTS adaptation. Calibration selected a common base threshold for all three rules at each target, allowing their report changes to be traced on the same instances.

\subsection{Calibration and frozen operating points}\label{sec:4.3}

Development calibration used risk targets \(\alpha \in \{ 0.05,0.10,0.20\}\), with 0.10 as the primary configuration. Base and THR thresholds used 200 equally spaced values from 0.50 to 0.999, plus an infinite endpoint. DELTA and DARTS used zero, 81 geometrically spaced values from \(10^{- 3}\) to \(10^{3}\), and a designated I-only endpoint. Complete-or-reject added zero to the threshold grid, giving 202 points. For each base-risk multiplier \(\kappa \in \{ 1,0.5\}\), an all-I calibration first selected a base threshold under target \(\kappa\alpha\). Each semantic rule was then calibrated on that support under \(\alpha\), and the better of the two base configurations was retained.

Calibration maximized source-normalized correct-report utility among nonempty outputs satisfying the empirical risk target, with risk tolerance \(10^{- 12}\). Within a parameter grid, utility ties favored lower risk and then more sources with output. Base calibration next favored the higher threshold. Semantic calibration instead favored fewer upgrades from I, followed by the smaller parameter-grid index. Across base configurations, utility ties favored lower risk, more sources with output, fewer upgrades, smaller semantic index and then larger \(\kappa\). Within-path utility comparisons used summed source-normalized correct weight with tolerance \(10^{- 12}\). Across-base comparisons used \(Y_{\mathrm{info}}\) with tolerance \(10^{- 10}\). A search without a feasible nonempty point returned its designated endpoint and recorded the feasibility status; zero total output had undefined risk.

The selected base thresholds were 0.9488492462, 0.8610854271 and 0.5025075377 at targets 5\%, 10\% and 20\%, respectively. The corresponding THR parameters were 0.8460402010, 0.7582763819 and 0.5451356784; DELTA used 3.3496543916, 1 and 0.1496235656. DARTS used 4.7315125896, 1.6788040181 and 0.1778279410. Complete-or-reject selected all rejection at 5\%, and thresholds 0.7733216080 and 0.5626884422 at the remaining targets. Restricted-menu controls held the corresponding base threshold fixed and repeated semantic calibration over the same rule-specific grid.

The original reporting configurations were selected on V and frozen before their test evaluation. The complete-or-reject and restricted-menu comparisons were subsequent analyses of existing prediction caches, with their parameters again selected on V. All finite-grid paths were evaluated descriptively on these caches. Grid summaries based on observed outcomes are reported as post-hoc diagnostics; the main comparisons retain the V-selected operating points. Source-transfer analyses applied the original predictor, scorer and frozen parameters to val. The val sources were separate from training and parameter selection for this prediction chain and had prior exposure in the broader project.

\subsection{Evaluation and development sensitivity}\label{sec:4.4}

We used the unique mask matching in Section~\ref{sec:3.5} with IoU threshold 0.5 and report weights \(\frac{1}{3}\), \(\frac{2}{3}\) and 1. The principal outcomes were report utility \(Y_{\mathrm{info}}\) and weighted reporting risk \(R_{\mathrm{info}}\), accompanied by report and correct-projection counts. Yield is measured per 100 frames after source normalization; risk is reported as a percentage. We write Y and R in compact figures and tables. All differences were computed from unrounded values, with risk differences expressed in percentage points.

Video source was the unit for paired comparisons. The eight test videos contributed equally to yield despite their different lengths, while count columns summarized emitted instances. Direction counts across rules and targets summarize repeated comparisons of these eight sources using the same predictions. Development sensitivity analyses used leave-one-source-out evaluation over the five V videos, fitting parameters on four sources and evaluating the fifth. They examined stricter matching, alternative report weights and changed ordering. The first two recalibrated THR and DELTA within each setting; the ordering analysis held DELTA actions fixed and recomputed matching.

\section{Results}\label{sec:5}

\subsection{Retaining useful semantics beyond complete-or-reject outputs}\label{sec:5.1}

Graded reporting retained correct partial descriptions for instances removed by complete-or-reject output. At the primary 10\% development target, all graded policies returned the same 8,010 admitted instances and preserved 7,376 correct instrument projections (Table~\ref{tab:2}). DELTA achieved a report utility of 88.93 at an observed risk of 9.66\%. The matched complete-or-reject policy emitted 1,348 reports, with utility 26.80 and risk 11.30\%. Its threshold had been selected on V; these risk values were measured after the frozen configurations were applied to test.

\begin{table*}[pos=!t]
\caption{Output choices at the 10\% development calibration target on the test population.}\label{tab:2}
\centering
\begin{tabular*}{\linewidth}{@{\extracolsep{\fill}}@{}>{\raggedright\arraybackslash}p{63mm}rrrrr@{}}
\toprule
\textbf{Reporting policy} & \(Y_{\mathrm{info}}\) & \(R_{\mathrm{info}}\) \textbf{(\%)} & \textbf{Reports} & \textbf{Correct I} & \textbf{Complete IVT} \\
\midrule
All I, same base support & 45.63 & 4.78 & 8,010 & 7,376 & 0 \\
All IVT, same base support & 95.97 & 33.24 & 8,010 & 7,376 & 5,015 \\
Complete or reject & 26.80 & 11.30 & 1,348 & 1,332 & 1,201 \\
THR & 85.51 & 9.19 & 8,010 & 7,376 & 1,550 \\
DELTA & 88.93 & 9.66 & 8,010 & 7,376 & 1,455 \\
DARTS adaptation & 84.67 & 7.93 & 8,010 & 7,376 & 1,087 \\
\bottomrule
\end{tabular*}
\par\smallskip
{\sffamily\fontsize{8}{9.5}\selectfont\noindent All policies share the initial predictions, correctness scores and base admission threshold. All-I and all-IVT display fixed levels; the remaining rows use parameters selected on V. Complete-or-reject outputs are rematched after filtering. Correct I counts correct instrument projections, and Complete IVT counts jointly correct complete reports. Y is source-normalized weighted correct-report yield per 100 frames; R is observed weighted reporting risk.\par}
\end{table*}

The endpoint references show how these outputs allocate semantic detail. All-I retained every base instance and achieved utility 45.63 at risk 4.78\%, with verbs and targets left unreported. All-IVT produced 5,015 complete correct triplets and utility 95.97, while its observed risk reached 33.24\%. Graded reporting occupied an intermediate position, retaining all base instances and displaying finer semantics where the correctness scores supported them. THR and DARTS yielded 85.51 and 84.67 at risks of 9.19\% and 7.93\%, respectively. Their complete correct-triplet counts differed because the rules assigned different levels to the same fixed predictions.

\begin{figure}[pos=!t]
\centering
\includegraphics{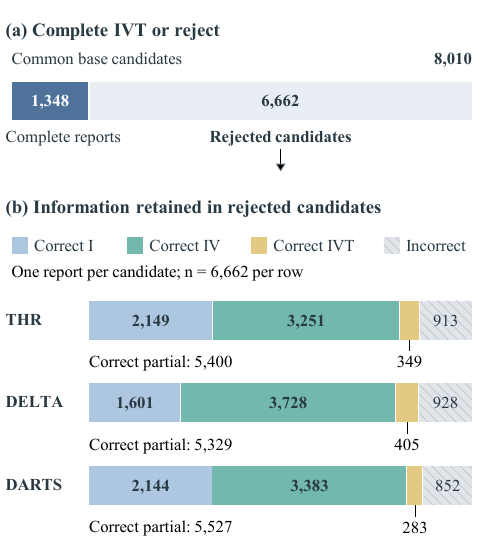}
\caption{Partial information retained among rejected candidates at the primary test configuration. (a) Complete-or-reject retains 1,348 of 8,010 base candidates. (b) Graded outputs on the remaining 6,662 candidates are partitioned by final action and joint correctness. The four categories are mutually exclusive, unweighted report counts. Correct I and correct IV denote complete correctness of those partial reports. Graded outputs retain their assignments from full-support matching; complete-or-reject uses its independently rematched retained set. The subset analysis uses the stored outputs of the frozen policies.}\label{fig:3}
\end{figure}

We next examined the 6,662 candidates rejected by the matched complete-or-reject policy (Fig.~\ref{fig:3}). On this subset, DELTA returned 1,601 correct I reports and 3,728 correct IV reports. Together, these 5,329 correct partial reports accounted for 80.0\% of the rejected candidates. DELTA also displayed 405 correct complete reports, while 928 reports were jointly incorrect. THR and DARTS retained 5,400 and 5,527 correct partial reports on the same subset. Each graded report retained its geometric assignment from the full base set, so this analysis traced information within the evaluated outputs.

The retained partial descriptions remained tied to the original instrument masks. Correct IV reports displayed an instrument and action that both agreed with the uniquely matched ground truth. Correct I reports preserved instrument identity when the policy selected the coarser level.

The other development targets changed the balance between partial descriptions and complete records. At 5\%, the specified complete-or-reject grid contained no feasible nonempty output on V, so its selected policy rejected all candidates. Its yield was zero and risk was undefined. At 20\%, complete-or-reject produced 3,917 complete correct IVT reports, compared with 3,802 for DELTA. Complete records and weighted report yield thus favored different output choices at particular working points.

\subsection{Utility of the intermediate instrument-verb level}\label{sec:5.2}

The intermediate IV level increased report utility for all three rules after each two-level control was separately recalibrated. At the primary target, both menus retained the same 8,010 reports and 7,376 correct instrument projections. Allowing IV increased DELTA utility from 65.80 to 88.93, a gain of 23.13 per 100 frames (Table~\ref{tab:3}). THR and DARTS gained 19.80 and 19.13, respectively. Both menus retained the same instances and used the same correctness scores, while their semantic parameters were calibrated separately. The gains therefore reflect the value of making IV available under each rule's calibrated selection.

\begin{table}[pos=!t]
\caption{Contribution of IV at the primary test configuration.}\label{tab:3}
\centering
\setlength{\tabcolsep}{3pt}
\begin{tabular*}{\linewidth}{@{\extracolsep{\fill}}@{}llrrr@{}}
\toprule
\textbf{Rule} & \textbf{Menu} & \textbf{Y} & \textbf{R (\%)} & \makecell[r]{\textbf{Complete}\\\textbf{IVT}} \\
\midrule
THR & I/IVT & 65.71 & 9.06 & 1,550 \\
THR & I/IV/IVT & 85.51 & 9.19 & 1,550 \\
DELTA & I/IVT & 65.80 & 8.66 & 1,508 \\
DELTA & I/IV/IVT & 88.93 & 9.66 & 1,455 \\
DARTS & I/IVT & 65.55 & 8.70 & 1,362 \\
DARTS & I/IV/IVT & 84.67 & 7.93 & 1,087 \\
\bottomrule
\end{tabular*}
\par\smallskip
{\sffamily\fontsize{8}{9.5}\selectfont\noindent Each I/IVT control was separately recalibrated on V using the same fixed base support and parameter grid as its three-level counterpart. All rows retain 8,010 reports and 7,376 correct I projections. Complete IVT counts jointly correct complete reports; DARTS denotes the adaptation in Section~\ref{sec:3.4}. Y is source-normalized weighted correct-report yield per 100 frames; R is weighted reporting risk.\par}
\end{table}

Adding IV changed risk and complete-triplet output differently across the three rules. THR changed risk by +0.12 percentage points and retained its 1,550 complete correct reports. DELTA increased risk by 1.01 points and returned 53 fewer complete correct reports. DARTS reduced risk by 0.77 points and returned 275 fewer complete correct reports. The available IV output therefore improved weighted correct-report yield in every pair, while each rule chose a different balance of detail and error.

\begin{figure}[pos=!t]
\centering
\includegraphics{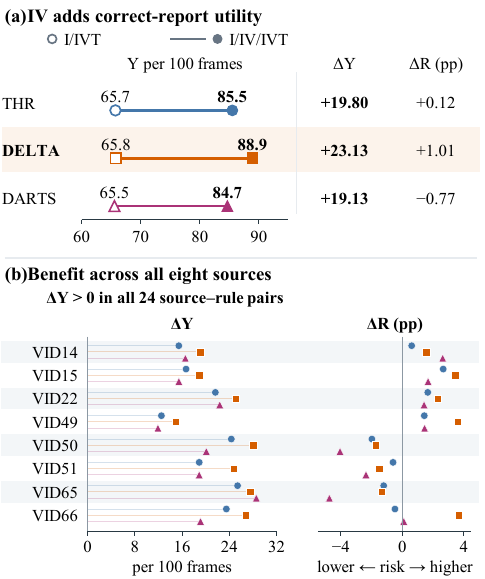}
\caption{Contribution of IV across eight test sources at the 10\% development target. (a) Aggregate utilities for recalibrated I/IVT controls and I/IV/IVT policies. (b) Source-level differences, computed as three levels minus two levels within each rule. All 24 source-rule utility differences are positive. These comparisons reuse eight sources and their fixed predictions. Risk differences are in percentage points, with positive values indicating greater observed risk; aggregate R uses source-normalized weights. Blue circles, orange squares and purple triangles identify THR, DELTA and DARTS, respectively.}\label{fig:4}
\end{figure}

The gain extended across all eight test sources (Fig.~\ref{fig:4}). Every source showed positive utility differences for each rule, giving 24 positive source-rule comparisons at the primary target. These source-level gains ranged from 11.90 to 28.57 per 100 frames. Risk differences ranged from -4.75 to +3.68 percentage points and varied in direction across sources. Thus the utility increase was distributed throughout the evaluated videos, including both long and short sequences. Source normalization gave each video the same contribution to the aggregate utility comparison.

\begin{figure}[pos=!t]
\centering
\includegraphics{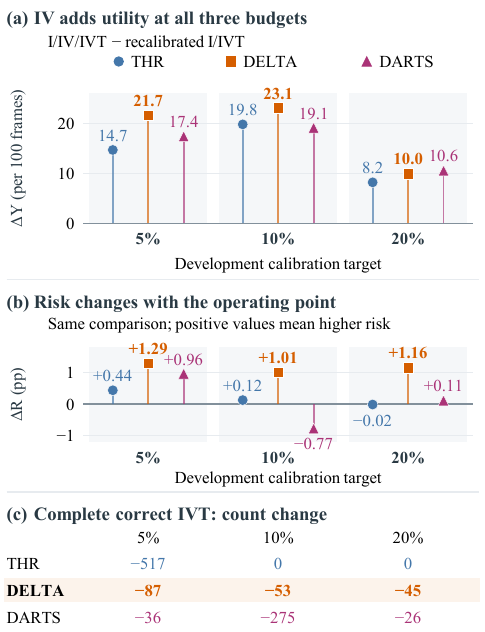}
\caption{Effect of IV across three development calibration targets. Differences compare I/IV/IVT with the corresponding I/IVT control, separately recalibrated on V. Panels show (a) report utility, (b) observed risk in percentage points and (c) complete correct-triplet counts. Base support is fixed within each pair and contains 6,835, 8,010 and 8,825 instances at targets 5\%, 10\% and 20\%, respectively. The targets represent separate configurations. Colors and shapes identify rules; the pale emphasis marks the primary target.}\label{fig:5}
\end{figure}

The two additional targets showed the same direction of utility gain (Fig.~\ref{fig:5}). Across the nine rule-target pairs, adding IV increased utility by 8.20 to 23.13 per 100 frames. At 5\%, DELTA gained 21.65, accompanied by a risk change of +1.29 percentage points and 87 fewer complete correct triplets. At 20\%, its gain was 9.96, with +1.16 points of risk and 45 fewer complete correct triplets. All 72 source-rule-target differences in utility were positive. The same eight video sources and predictions contributed to these repeated comparisons.

The three targets admitted 6,835, 8,010 and 8,825 instances, respectively. Each menu comparison therefore held support fixed within its target, while different targets evaluated different admitted sets. IV increased utility at every tested configuration, with the size of the gain and complete-record output varying by rule and working point.

\subsection{Operating-point dependence of risk and utility}\label{sec:5.3}

\begin{figure*}[pos=!t]
\centering
\includegraphics{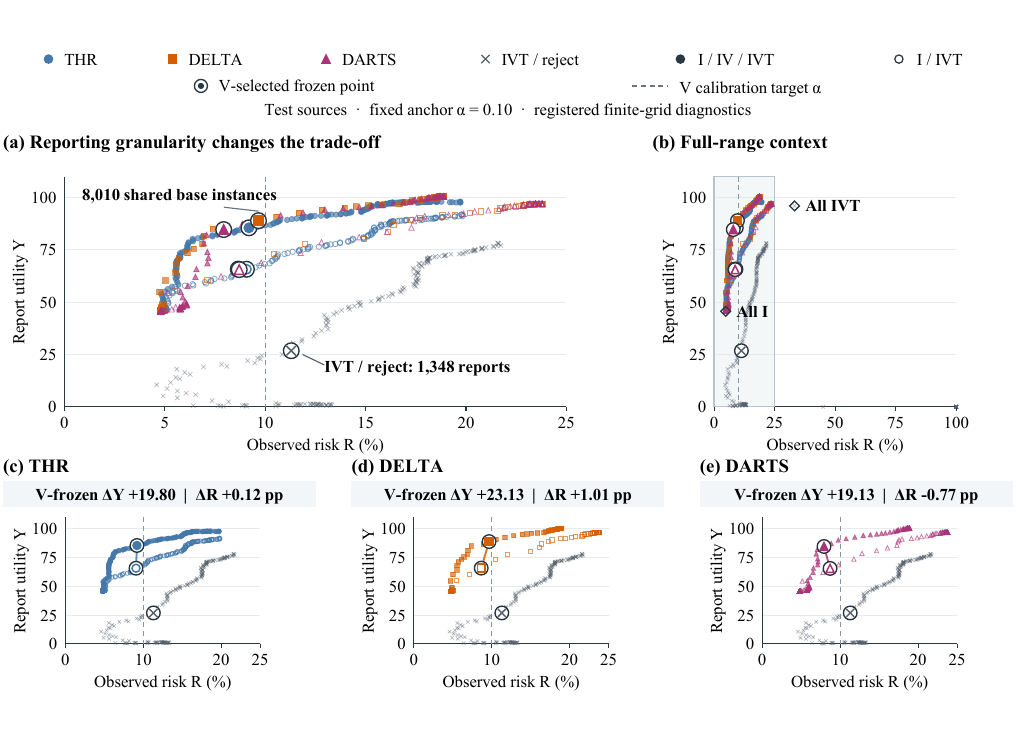}
\caption{Observed risk and report utility over the finite grids on the primary test base set. (a) Detail over 0-25\% observed risk; (b) the full 0-100\% range, including fixed-level references; (c-e) the individual rule paths. Filled and open markers denote three-level and two-level menus, respectively; crosses denote complete-or-reject. Enlarged outlined markers are the V-selected frozen points. The figure evaluates 734 hierarchy-grid records and 202 complete-or-reject records, displaying points with defined risk inside each window. The dashed line marks the 10\% development target. Paths summarize post-hoc evaluation of stored predictions; zero-output risk remains undefined. Y is source-normalized weighted correct-report yield per 100 frames and can exceed 100; R is weighted reporting risk.}\label{fig:6}
\end{figure*}

The finite parameter paths showed that reporting-rule comparisons depended on the observed-risk range. Figure~\ref{fig:6} holds the primary base set fixed and evaluates the three-level and two-level menus over their specified grids. Enlarged markers identify the operating points selected on V. At the V-selected points, three-level outputs retained greater utility than their restricted-menu counterparts, while complete-or-reject returned substantially fewer reports. Its selected point contained 1,348 reports compared with 8,010 for every instance-preserving configuration.

The fixed-level endpoints help interpret the full range. All-I had risk 4.78\% and utility 45.63, whereas all-IVT reached risk 33.24\% and utility 95.97. Between these references, changing a semantic parameter altered how many selected instances received partial or complete descriptions. The three-level paths could retain correct IV content while reducing exposure to unreliable targets. The complete-or-reject path changed both semantic output and the retained set, with ground-truth assignments recomputed for each threshold.

The original V-selected points placed DELTA at utility 88.93 and DARTS at 84.67, with different observed risks. A post-hoc query of the finite test grid under \(R_{\mathrm{info}} \leq 10\%\) instead found maximum utilities of 90.2821 for DARTS and 88.9312 for DELTA. This query summarized grid points using their observed test outcomes; the V-selected configurations remained unchanged. The result separates the recurring benefit of the IV menu from the choice of a particular decision rule.

The same development target produced different test risks in Table~\ref{tab:2}. Calibration selected discrete configurations on V; their test risks depended on the evaluated instances and displayed fields.

\subsection{How report refinements change semantic content}\label{sec:5.4}

\begin{figure}[pos=!t]
\centering
\includegraphics{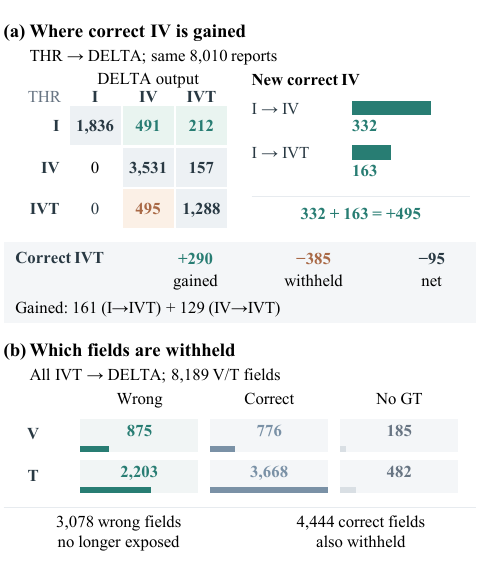}
\caption{Semantic changes on the primary test support. (a) THR-to-DELTA transitions over 8,010 reports, with the origins of 495 newly displayed correct IV projections and the gains and losses of complete correct IVT. The separate count of 495 IVT-to-IV transitions includes 385 fully correct triplets. Joint report correctness changes by 73 gains and 60 losses. Projection, complete-triplet and joint-correctness events overlap. (b) Fields withheld by DELTA relative to all-IVT, classified against the shared ground-truth assignment. Counts refer to fields; No GT denotes an absent unique match.}\label{fig:7}
\end{figure}

At the primary configuration, DELTA displayed 495 more correct IV projections than THR on the same 8,010 instances (Fig.~\ref{fig:7}a). Of these newly displayed correct pairs, 332 came from I-to-IV transitions and 163 from I-to-IVT transitions. No previously displayed correct IV pair was removed. The projection count includes correct I/V pairs within IVT reports, including those with an incorrect target. These gains added correct action descriptions to already retained instrument reports.

Complete correct-triplet output changed in both directions. DELTA gained 290 correct complete reports: 161 from I-to-IVT transitions and 129 from IV-to-IVT transitions. It also withheld the target in 385 previously correct complete reports, producing a net change of -95. Those reports continued to display their correct IV pair. The action matrix contained 495 IVT-to-IV transitions in total; 385 involved fully correct predictions. Jointly correct reports increased by 13, from 73 gains and 60 losses, while report weights changed with granularity.

A second comparison traced which fields DELTA withheld relative to all-IVT (Fig.~\ref{fig:7}b). It omitted 8,189 verb or target fields: 3,078 were wrong, 4,444 were correct and 667 belonged to candidates without a unique ground-truth match. Targets accounted for 6,353 omitted fields, including 2,203 wrong labels and 3,668 correct labels. The remaining 1,836 omissions concerned verbs. Withholding fields reduced erroneous statements and left some correct content unreported. Unmatched reports contributed errors under the joint report criterion in Section~\ref{sec:3.5}.

\begin{figure*}[pos=!t]
\centering
\includegraphics{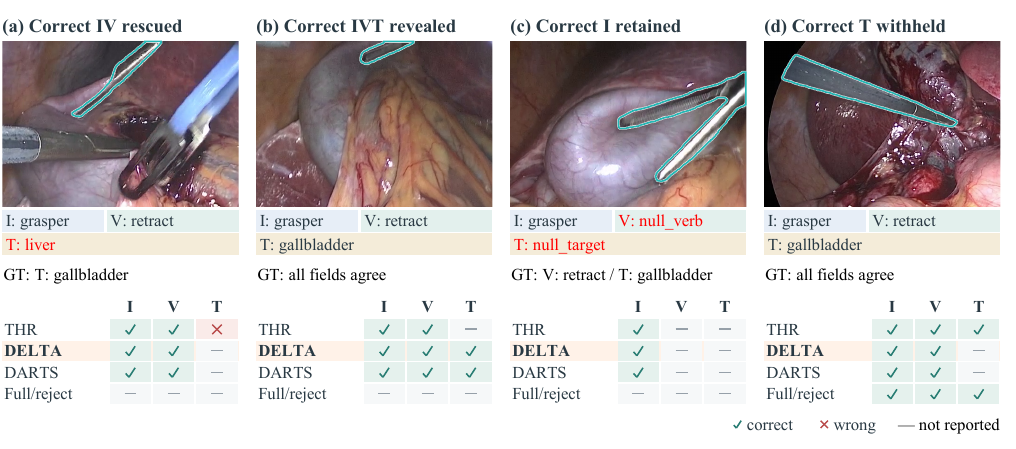}
\caption{Representative report decisions at the primary frozen configuration. (a) Correct IV retained despite an incorrect target; (b) a correct complete triplet revealed by DELTA and DARTS; (c) correct instrument identity retained when verb and target predictions are wrong; (d) a correct target withheld by DELTA and DARTS. Cyan contours mark predicted instrument masks. Masks and label values remain fixed within each case, and correctness symbols use ground truth after inference. Full/reject denotes the complete-or-reject policy and uses its actual rematched retained output. Horizontal strokes denote omitted fields; null\_verb and null\_target are class labels. Panel (c) revisits the instance in Fig.~\ref{fig:1}(b).}\label{fig:8}
\end{figure*}

Figure~\ref{fig:8} shows the corresponding choices on four representative instances. In panel (a), the predicted liver target was wrong, while the grasper and retract labels were correct. DELTA and DARTS displayed the correct IV pair; THR displayed the incorrect full triplet, and complete-or-reject removed the instance. In panel (b), the same two rules revealed a correct gallbladder target that THR left unreported. Complete-or-reject also removed this instance, illustrating how different calibrated rules can retain different complete predictions.

Panel (c) shows the coarser endpoint of this behavior: all three graded rules retained the correct instrument and omitted the incorrect verb and target. Panel (d) illustrates the associated completeness cost. Its predicted triplet was correct, but DELTA and DARTS displayed IV, whereas THR and complete-or-reject retained the complete record.

\subsection{Source transfer and development sensitivity}\label{sec:5.5}

Applying the same frozen configurations to the later val sources increased observed reporting risk across all nine rule-target settings (Table~\ref{tab:4}). At the primary target, risks were 16.38\% for THR, 17.82\% for DELTA and 17.80\% for DARTS. Across all targets, increases relative to test ranged from 5.34 to 9.87 percentage points, and every val risk exceeded its development target. The predictor, scorer and rule parameters were unchanged. These measurements establish the source dependence of the realized error level and motivate checking calibration when the reporting population changes.

\begin{table}[pos=!t]
\caption{Observed risk under transfer of the same frozen configurations.}\label{tab:4}
\centering
\setlength{\tabcolsep}{3pt}
\begin{tabular*}{\linewidth}{@{\extracolsep{\fill}}@{}llrrr@{}}
\toprule
\textbf{Target} & \textbf{Rule} & \makecell[r]{\textbf{Test R}\\\textbf{(\%)}} & \makecell[r]{\textbf{Val R}\\\textbf{(\%)}} & \(\Delta R\) \textbf{(pp)} \\
\midrule
5\% & THR & 4.09 & 9.42 & +5.34 \\
5\% & DELTA & 3.92 & 10.45 & +6.52 \\
5\% & DARTS & 3.84 & 9.46 & +5.61 \\
10\% & THR & 9.19 & 16.38 & +7.19 \\
10\% & DELTA & 9.66 & 17.82 & +8.15 \\
10\% & DARTS & 7.93 & 17.80 & +9.87 \\
20\% & THR & 20.08 & 26.24 & +6.16 \\
20\% & DELTA & 20.38 & 27.17 & +6.78 \\
20\% & DARTS & 20.57 & 26.48 & +5.91 \\
\bottomrule
\end{tabular*}
\par\smallskip
{\sffamily\fontsize{8}{9.5}\selectfont\noindent Val contains 2,443 frames from three later-evaluated video sources. The original predictor, correctness scorer and V-selected parameters are fixed in both populations. Differences are val minus test, calculated before rounding. Target denotes the development calibration target; R denotes weighted reporting risk, and pp denotes percentage points.\par}
\end{table}

Development sensitivity analyses examined matching, report weighting and candidate ordering on five V sources (Table~\ref{tab:5}). With matching tightened to IoU 0.75, recalibrated DELTA exceeded THR by 0.77 in utility at the 10\% target, with 0.73 percentage points lower risk. We also replaced field-fraction weights with \(w_{a} = \log(100/n_{a})/\log(100)\), where \(n_{a}\) is the number of legal triplets compatible with the reported fields. Under these weights, DELTA gained 2.42 in utility and reduced risk by 0.95 points. The two settings used leave-one-source-out calibration for the THR and DELTA comparison.

\begin{table*}[pos=!t]
\caption{Development sensitivity at the 10\% target using five-source leave-one-source-out evaluation on V.}\label{tab:5}
\centering
\begin{tabular*}{\linewidth}{@{\extracolsep{\fill}}@{}>{\raggedright\arraybackslash}p{43mm}>{\raggedright\arraybackslash}p{73mm}rr@{}}
\toprule
\textbf{Setting} & \textbf{Comparison} & \(\Delta Y\) & \(\Delta R\) \textbf{(pp)} \\
\midrule
IoU = 0.75 & DELTA - THR & +0.77 & -0.73 \\
Compatible-set weights & DELTA - THR & +2.42 & -0.95 \\
Report-score ordering & Selected score - base I score (DELTA) & -0.01 & +0.01 \\
\bottomrule
\end{tabular*}
\par\smallskip
{\sffamily\fontsize{8}{9.5}\selectfont\noindent \(\Delta Y\) is the change in source-normalized weighted correct-report yield per 100 frames; \(\Delta R\) is the change in weighted reporting risk, in percentage points (pp). IoU denotes intersection over union. IoU and compatible-set-weight rows compare recalibrated DELTA minus THR. The ordering row holds DELTA actions fixed and compares ordering by the selected-action score with ordering by the base instrument score, rematching in each order. Its changes are -0.01319 in Y and +0.01279 percentage points in R, shown to five decimal places.\par}
\end{table*}

Changing ordering had a much smaller effect when DELTA actions were fixed. Sorting by the selected-action score changed utility by -0.01319 and risk by +0.01279 percentage points relative to base-score ordering. At the 5\% target, both the stricter-IoU and alternative-weight settings placed DELTA at lower utility and lower risk than THR.

\section{Discussion}\label{sec:6}

Semantic granularity is a useful design variable for reporting surgical instance predictions. Once an instrument instance has been admitted, its identity can remain available while the report selects how much action and target detail to display. This separation gives partial predictions an explicit output form and a joint correctness criterion tied to their instrument masks. At the primary test configuration, DELTA retained correct partial descriptions for 80.0\% of the candidates rejected by the matched complete-or-reject policy (Fig.~\ref{fig:3}). This connects selective and hierarchical prediction (Section~\ref{sec:2}) to reports whose semantic correctness remains tied to a specific spatial instance.

The intermediate IV level provides a useful option when the evidence supports an instrument-action pair more strongly than its complete triplet. Joint correctness requires every displayed field to agree with the same matched instance, so adding a target introduces another condition. An IV report can retain a correct action description when that condition fails, as illustrated in Fig.~\ref{fig:8}(a). Utility increased for every tested rule-target pair and corresponding source comparison after the restricted controls were recalibrated (Table~\ref{tab:3}; Figs.~\ref{fig:4} and~\ref{fig:5}). With masks, labels, scores and base support fixed, these gains arise from the reporting menu and its calibrated selection. The menu supplies an intermediate action description; the correctness scores and decision rule determine when that description is selected.

The value of a partial report also depends on what its recipient needs to recover. Field-fraction weights give a correct IV description more credit than a correct I description, while reserving full credit for a correct triplet. The field analysis makes the associated completeness cost concrete. Relative to all-IVT output, DELTA withheld 3,078 incorrect fields and 4,444 correct fields at the primary configuration (Fig.~\ref{fig:7}b). Coarsening can therefore remove an erroneous statement or leave a correct detail unreported, as the examples in Fig.~\ref{fig:8} show. A task that can use instrument-action descriptions may benefit from retaining them; a task requiring complete records must also account for omitted targets. The admitted candidates, complete labels, correctness scores and rule parameters specify each reporting decision. A recipient with these inputs and the base ordering can reproduce the selection.

Choosing the available report levels and choosing a decision rule address different parts of the design. The IV option increased utility across THR, DELTA and the DARTS adaptation, while their operating points differed in observed risk and complete-triplet yield. At the three primary V-selected operating points, DELTA had the highest utility and risk. In the post-hoc finite-grid comparison under a 10\% observed test-risk ceiling, DARTS achieved the greatest utility (Fig.~\ref{fig:6}). These comparisons make the intended operating conditions central to rule selection. Report utility, weighted reporting risk and complete correct-triplet output describe complementary consequences of the selection. Tracking instance support alongside them identifies how much of the change comes from retaining instances or changing their descriptions.

Source transfer makes calibration part of the reuse decision \citep{ref39}. With the predictor, scorer and reporting parameters fixed, observed reporting risk exceeded the development target in all nine val configurations (Table~\ref{tab:4}). At the primary 10\% target, observed risks ranged from 16.38\% to 17.82\%. The realized error level depends on the evaluated population, including the admitted instances and the correctness of their displayed fields. Applying the framework to another population therefore calls for target-source evaluation of both instance admission and semantic selection \citep{ref40}. The CholecTriplet-Seg evaluation provides this assessment for fixed Mask2Former predictions. The development analyses in Table~\ref{tab:5} further identify matching and utility weighting as explicit choices to examine when adapting the reporting objective.

Comparing correctness scores across additional predictors and prospectively reserved sources would clarify which score properties sustain the IV level's contribution to report utility. Holding the reporting menu fixed would allow these comparisons to examine how source adaptation changes reporting outcomes. A complementary study could estimate report weights from a specified downstream task, including the costs of wrong displayed fields and omitted correct fields. Comparing these weights with the field-fraction objective would connect report selection to that task's measured requirements.

\section{Conclusion}\label{sec:7}

We introduced instance-preserving graded semantic reporting for surgical action triplets, separating instance admission from the semantic detail displayed for each retained mask. The framework selects I, IV or IVT using hierarchical correctness scores while preserving predicted masks, displayed label values and base ordering. On fixed CholecTriplet-Seg predictions, matched complete-or-reject comparisons showed that graded outputs retain correct partial descriptions among rejected candidates. Separately recalibrated two-level controls established the contribution of IV, with higher report utility across all tested rules and development targets. Instance-level analyses traced these gains to changes in displayed semantics and their trade-off with complete correct-triplet output. Choosing a reporting policy requires joint consideration of utility, observed risk and complete-record requirements on the intended source. Keeping partial descriptions attached to their instruments makes it possible to assess semantic yield and completeness on a common set of instances.

\section*{Funding}
The authors received no specific funding for this work.

\section*{Declaration of competing interest}
The authors declare no conflict of interest.

\section*{Data availability}
The CholecTriplet-Seg dataset used in this study \citep{ref03} is available from Synapse at \url{https://doi.org/10.7303/syn70783776.1}.

\section*{Code availability}
The code supporting this study are being prepared for public release at \url{https://github.com}

\section*{Declaration of generative AI and AI-assisted technologies in the manuscript preparation process}
During the preparation of this work, the authors used ChatGPT (OpenAI) to assist with language editing and improve the readability of the manuscript. The authors reviewed and edited the resulting content as needed and take full responsibility for the content of the article.


\end{document}